\documentclass[]{jingdong}
\usepackage[toc,page,header]{appendix}
\usepackage{latexsym}
\usepackage[T1]{fontenc}
\usepackage[utf8]{inputenc}
\usepackage{microtype}
\usepackage{inconsolata}
\usepackage{graphicx}
\usepackage{hyperref}       
\usepackage{url}            
\usepackage{booktabs}       
\usepackage{amsfonts}       
\usepackage{nicefrac}       
\usepackage{stackengine}
\usepackage{microtype}      
\usepackage{colortbl}
\usepackage{xcolor}
\definecolor{Green}{rgb}{0,1.0,0}
\definecolor{Red}{rgb}{1.0,0,0}
\usepackage{pifont}
\newcommand{\xmark}{\ding{55}} 
\newcommand{\cmark}{\ding{51}} 
\usepackage{amsmath}
\usepackage{amssymb}
\usepackage{amsthm}
\usepackage{mathrsfs}
\usepackage{pifont}
\usepackage{MnSymbol}
\usepackage{balance}
\usepackage{enumitem}
\usepackage{listings}
\usepackage{xcolor}
\usepackage{natbib}
\usepackage{multicol}
\usepackage{subcaption}

\AtBeginDocument{%
  \providecommand\BibTeX{{%
    \normalfont B\kern-0.5em{\scshape i\kern-0.25em b}\kern-0.8em\TeX}}}

\makeatletter
\DeclareRobustCommand\onedot{\futurelet\@let@token\@onedot}
\def\@onedot{\ifx\@let@token.\else.\null\fi}

\usepackage{setspace}
\usepackage{mathtools}

\usepackage{multirow,booktabs}
\usepackage{subcaption}

\newcommand{\owo}[1]{\textsc{OAgents}}

\definecolor{lightgreen}{RGB}{144, 238, 144} 
\definecolor{lightred}{RGB}{255, 105, 97}

\newtcolorbox{promptbox}[2][Prompt]{
colback=black!5!white,
arc=5pt, 
boxrule=0.5pt,
fonttitle=\bfseries,
title=#1, 
before upper={\small}, fontupper=\fontfamily{ptm}\selectfont,
colframe=#2, 
}
\definecolor{ogreen}{RGB}{34, 139, 34}

\usepackage{cleveref}
\theoremstyle{plain}

\theoremstyle{definition}

\theoremstyle{remark}
\usepackage{subcaption}

\usepackage[textsize=tiny]{todonotes}
\usepackage{fontawesome}

\title{EgoLive: A Large-Scale Egocentric Dataset from Real-World Human Tasks}

\affiliation{Joy Future Academy, JD}

\abstract{ 
The advancement of robot learning is currently hindered by the scarcity of large-scale, high-quality datasets. While established data collection methods such as teleoperation and universal manipulation interfaces dominate current datasets, they suffer from inherent limitations in scalability and real-world deployability. Human egocentric video collection, by contrast, has emerged as a promising approach to enable scalable, natural and in-the-wild data collection.
As such, we present \textbf{EgoLive}, a large-scale, high-quality egocentric dataset designed explicitly for robot manipulation learning. \textbf{EgoLive} establishes three distinctive technical advantages over existing egocentric datasets: first, it represents the largest open-source annotated egocentric dataset focused on real-world task-oriented human routines to date; second, it delivers leading data quality via a customized head-mounted capture device and comprehensive high-precision multi-modal annotations; third, all data is collected exclusively in unconstrained real-world scenarios and encompasses vertical field human working data, including home service, retail, and other practical work scenarios, providing superior diversity and ecological validity.
With the introduction of \textbf{EgoLive}, we aim to provide the research community with a scalable, high-quality dataset that accelerates breakthroughs in generalizable robotic models and facilitates the real-world deployment of robot systems.

}

\checkdata[Dataset]{\url{https://robotdata-market.jdcloud.com/console/market}}
\begin{document}

\maketitle
\section{Introduction}

The lack of large-scale datasets has emerged as a critical bottleneck in advancing the generalization capabilities of contemporary robot manipulation learning models. While recent progress in pre-training Vision-Language-Action (VLA) models \cite{physical_intelligence_2024, gr00tn1_2025} has demonstrated remarkable progress in facilitating real-world robot manipulation, the pursuit of enhanced generalization cannot be achieved solely through scaling existing robot datasets. Existing datasets are mainly constrained by two key limitations: one is restricted environmental diversity that fails to capture real-world complexity; the other is poor extensibility that makes large-scale data collection implausible. These constraints collectively hinder the development of truly generalizable embodied systems capable of operating in open-world scenarios. 

Egocentric data \cite{Egodex, egocentric10k, xperience_10m} present a promising solution to address current limitations in acquiring data at scale for robot learning. By capturing first-person visual perspectives using head-mounted wearable devices, this approach embodies human-like visual perception in a natural manner and offers several key advantages. \textbf{Human Learning Inspiration:} The egocentric approach advocates human-centric visual perception and learning processes, enabling curiosity-driven exploration similar to infant developments \cite{YU2012,egocentric_info_infant_learning} and supporting incremental skill acquisition through cumulative first-person experience. \textbf{In-the-wild Collection:} Unlike fixed camera systems, this method allows for unrestricted movement in real-world environments, capturing natural human behaviors and interactions without spatial constraints or significant physical interference on the wearer. \textbf{Scalability Potential:} The relatively simple hardware setup (e.g., single head-mounted camera) enables large-scale data collection across diverse environments and users with low equipment costs. 

\begin{figure}[t!]
    \centering
    \includegraphics[width=1\linewidth]{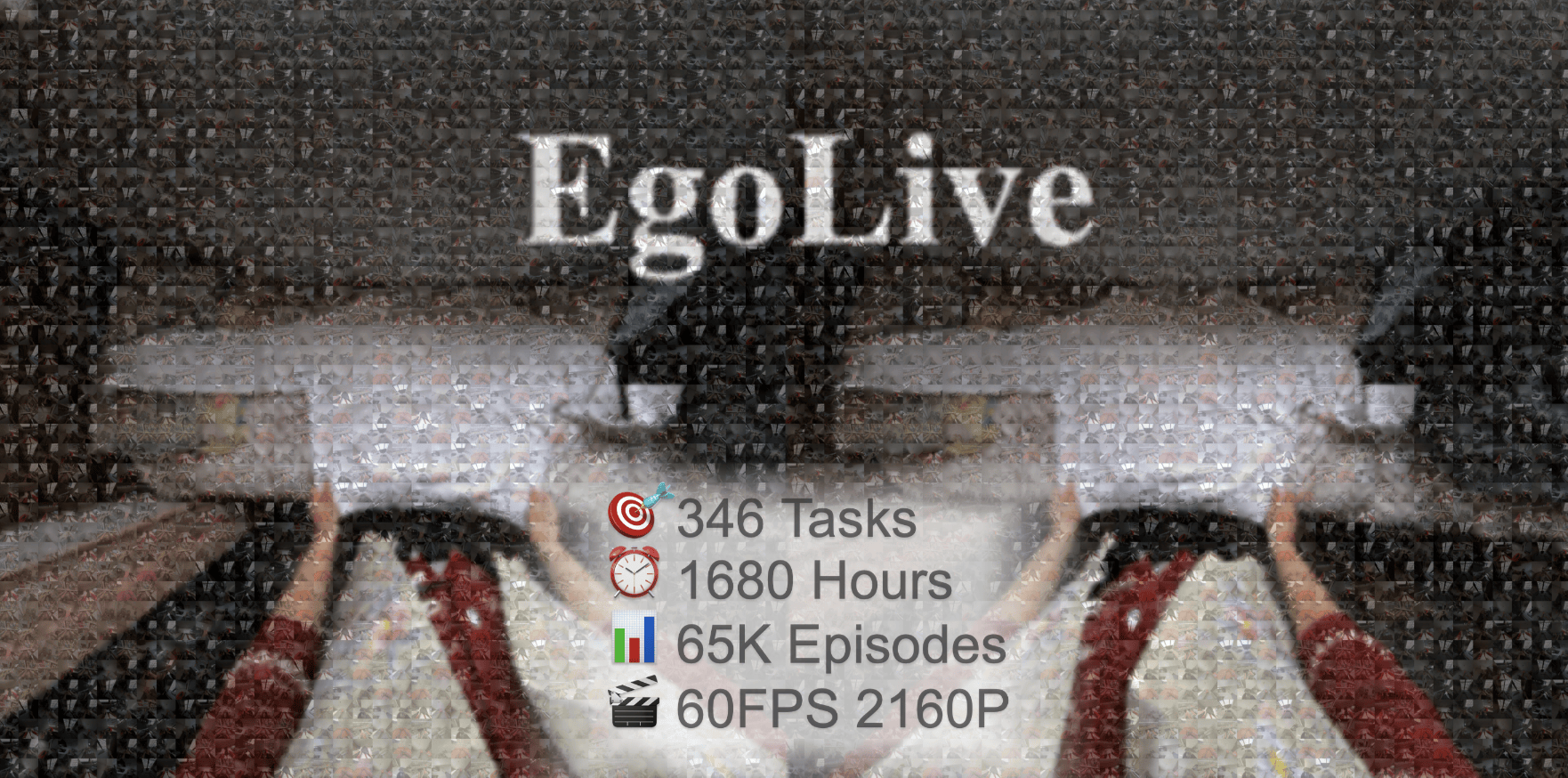}
    \caption{EgoLive is a large-scale high-quality and diverse egocentric dataset. It records human daily manipulation tasks with stereo vision.}
    \label{fig:cover}
\end{figure}

We introduce EgoLive, a large-scale egocentric dataset capturing real-world human task demonstrations. It comprises $1,680$ hours of high-fidelity stereo video recordings at $60$ frames per second, with $65,866$ episodes spanning $346$ different real-world tasks. Notably, EgoLive exhibits several key characteristics that distinguish it from existing egocentric datasets:

\begin{itemize}
    \item To our knowledge, EgoLive is the largest open-source annotated egocentric dataset for real-world human tasks to date.
    \item The dataset achieves superior quality in two dimensions. \textbf{Collection specifications}:  Our custom-designed head-mounted capture device achieves human-like stereo vision with a $130^{\circ}$ horizontal × $130^{\circ}$ vertical field of view (FOV), capturing video at 60 frames per second with $2160$×$2160$ resolution per camera. \textbf{Annotation richness and accuracy}: It provides comprehensive multi-modal annotations, including $6$-DoF motion tracking, fine-grained semantic segmentation and 3D scene reconstruction, all with high accuracy.
    \item In contrast to existing egocentric datasets, EgoLive is collected entirely in unconstrained real-world environments, offering significantly improved scene diversity and ecological validity for robot learning.
\end{itemize}

\section{Related Work}
\label{sec:related}

\subsection{Manipulation Data Collection}
Manipulation data collection falls broadly into three paradigms: \textbf{real-robot teleoperation}, \textbf{universal manipulation interfaces} (UMI), and \textbf{human egocentric video collection}.

\textbf{Real-robot teleoperation} systems primarily employ three approaches: isomorphic master-follower control, VR-based systems, and motion capture. Notable implementations include ALOHA \cite{ALOHA} (low-cost synchronized teleoperation via tethered robotic arms), AgiBot \cite{agibot_2024} (VR/motion-capture humanoid control), and Tesla Optimus \cite{tesla_optimus_2024} (wearable motion capture suits). While these methods generate high-fidelity trajectories with strong physical priors for policy learning, they face scalability limitations due to specialized hardware requirements, intensive human involvement, and prohibitive costs for large-scale data collection.

\textbf{Universal Manipulation Interfaces} (UMI) \cite{UMI} employ instrumented grippers with GoPro cameras to capture human demonstrations, aligning human-robot perceptual views. FastUMI \cite{Fastumi} enhances tracking precision via optimized wrist pose estimation. Recent works like Gen series \cite{Gen-0,Gen-1} leverage UMI data for scalable manipulation policy training. Although these handheld interfaces enable natural interaction and reduce visual domain gaps, they are typically tailored to specific robot embodiments and thus offer limited cross-platform compatibility. In addition, UMI systems remain difficult to adapt to dexterous hands, further limiting their application scope.

\textbf{Human egocentric video collection} offers a scalable, minimal-hardware paradigm for capturing natural interactions through wearable cameras. EgoScale \cite{Egoscale} extracts high-DoF dexterous hand motions from $20$k$+$ hours of in-the-wild videos. EgoDex \cite{Egodex} enables large-scale bare-hand tracking via head-mounted devices; EgoMimic \cite{Egomimic} implements portable multi-modal data collection through smart glasses. Compared to real-robot teleoperation, this vision-only approach eliminates hardware constraints and spatial limitations, significantly accelerating real-world data acquisition. Unlike physical gripper-based interfaces like UMI, it preserves natural human hand morphology and thereby supports direct collection of high-DoF dexterous manipulation data.

\subsection{Human Egocentric Dataset}

\begin{table}[b!]
    \centering
    \renewcommand{\arraystretch}{1.12}
    \resizebox{\textwidth}{!}{
    \begin{tabular}{lcccccccc}
        \toprule
        \textbf{Dataset} & \textbf{Scene} & \textbf{Scale} & \textbf{Resolution} & \textbf{FPS} & \textbf{Multi-view} & \textbf{Motion-tracking} & \textbf{Language Annot.} & \textbf{Depth} \\
        \midrule

        \multicolumn{9}{l}{\textit{Generalist}} \\
        \textbf{EPIC-KITCHENS-100} \cite{damen2022rescaling}
        & Real-world
        & 100 h
        & 1920*1080
        & 50
        & \textcolor{Red}{\xmark}
        & \textcolor{Red}{\xmark}
        & \textcolor{Green}{\cmark}
        & \textcolor{Red}{\xmark} \\

        \textbf{Ego4D} \cite{grauman2022ego4d}
        & Real-world
        & 3680 h
        & Inconsistent
        & 30
        & \textcolor{Red}{\xmark}
        & \textcolor{Red}{\xmark}
        & \textcolor{Green}{\cmark}
        & \textcolor{Red}{\xmark} \\

        \midrule
        \multicolumn{9}{l}{\textit{Manipulation-Centric}} \\
        \textbf{EgoMimic} \cite{Egomimic}
        & Laboratory
        & 55 h
        & 1408*1408
        & 30
        & \textcolor{Red}{\xmark}
        & \textcolor{Green}{\cmark}
        & \textcolor{Red}{\xmark}
        & \textcolor{Red}{\xmark} \\

        \textbf{HOI4D} \cite{liu2022hoi4d}
        & Laboratory
        & 22 h
        & 1920*1080
        & 15
        & \textcolor{Red}{\xmark}
        & \textcolor{Green}{\cmark}
        & \textcolor{Red}{\xmark}
        & \textcolor{Green}{\cmark} \\

        \textbf{HOT3D} \cite{banerjee2025hot3d}
        & Laboratory
        & 14 h
        & 1408*1408
        & 30
        & \textcolor{Green}{\cmark}
        & \textcolor{Green}{\cmark}
        & \textcolor{Red}{\xmark}
        & \textcolor{Red}{\xmark} \\

        \textbf{EgoDex} \cite{Egodex}
        & Laboratory
        & 829 h
        & 1920*1080
        & 30
        & \textcolor{Red}{\xmark}
        & \textcolor{Green}{\cmark}
        & \textcolor{Green}{\cmark}
        & \textcolor{Red}{\xmark} \\

        \midrule
        \multicolumn{9}{l}{\textit{Deployment-Scale}} \\
        \textbf{Egocentric-10K} \cite{egocentric10k}
        & Real-world
        & 10k h
        & 1920*1080
        & 30
        & \textcolor{Red}{\xmark}
        & \textcolor{Red}{\xmark}
        & \textcolor{Red}{\xmark}
        & \textcolor{Red}{\xmark} \\

        \textbf{Xperience-10M} \cite{xperience_10m}
        & Real-world
        & 1059 h
        & 512*512
        & 20
        & \textcolor{Green}{\cmark}
        & \textcolor{Green}{\cmark}
        & \textcolor{Green}{\cmark}
        & \textcolor{Green}{\cmark} \\

        \midrule
        \textbf{EgoLive (Ours)}
        & \textbf{Real-world}
        & 1680 h
        & \textbf{2160*2160}
        & \textbf{60}
        & \textcolor{Green}{\cmark}
        & \textcolor{Green}{\cmark}
        & \textcolor{Green}{\cmark}
        & \textcolor{Green}{\cmark} \\
        \bottomrule
    \end{tabular}
    }
    \caption{Comparison of representative human egocentric datasets relevant to embodied manipulation and human-to-robot transfer. A modality is marked as present only if provided in the main public release of the dataset. EgoLive targets real-world scenes with the second longest collection duration and achieves superior spatiotemporal resolution and annotation comprehensiveness.}
    \label{tab:egocentric_dataset_comparison}
    \vspace{-0.3cm}
\end{table}

Existing human egocentric datasets can be categorized into three major trends: \textbf{Generalist}, \textbf{Manipulation-centric}, and \textbf{Deployment-scale}. They reflect different design priorities for first-person data collection, namely semantic breadth, interaction fidelity, and real-world operational coverage.

\textbf{Generalist} related works such as EPIC-KITCHENS-100 \cite{damen2022rescaling} and Ego4D \cite{grauman2022ego4d} have played a foundational role in egocentric visual understanding. They emphasize large-scale video coverage, rich language supervision, and wide diversity of daily activities and environments. As a result, they have been highly influential for action recognition and general-purpose first-person representation learning. However, these datasets do not offer a unified set of manipulation-oriented geometric signals such as camera trajectories, 3D hand keypoints, dense hand/object masks, and procedural sub-task boundaries. This limits their direct use for learning fine-grained manipulation policies or structured operational procedures without relying on additional resources beyond the core release.

\textbf{Manipulation-centric} works move closer to embodied skill learning by making hand-object interaction directly observable. EgoMimic \cite{Egomimic} shows the value of pairing egocentric video with camera motion and 3D hand tracking for robot learning. HOI4D \cite{liu2022hoi4d} provides substantially denser supervision, including RGB-D, camera pose, 3D hand pose, segmentation masks and action segmentation, making it particularly useful for category-level hand-object interaction analysis. HOT3D \cite{banerjee2025hot3d} strengthens the 3D tracking perspective through accurate hand, object, and camera pose annotations in an egocentric multi-view setup. EgoDex \cite{Egodex} advances this paradigm both in volume and annotation richness, providing large-scale dexterous tabletop manipulation data that combines language annotations with camera pose and native 3D hand tracking. Collectively, these datasets are valuable for learning manipulation priors, action grounding, and demonstration-based transfer. Nevertheless, most of them remain centered on household, laboratory, or tabletop settings and therefore provide limited coverage of longer-horizon workflows in real service or business operations.

\textbf{Deployment-scale} related works extend egocentric data collection toward greater operational diversity and more realistic long-duration capture. Egocentric-10K \cite{egocentric10k} introduces large-scale first-person recording to real factory environments, improving coverage of real operational footage but providing only relatively sparse annotation types. Xperience-10M \cite{xperience_10m} substantially expands its multi-modal sensory and annotation profile, combining synchronized first-person video with depth, camera pose, hand motion capture, and hierarchical language annotations that encompass task and sub-task structures. These datasets represent important steps toward deployment-oriented embodied intelligence. 

\textbf{EgoLive} pushes the boundary further by collecting human egocentric data in real service-oriented business scenarios including home services, retail and pharmacy which remain largely underexplored in existing egocentric datasets. In addition to RGB videos and language annotations, this dataset includes camera pose, 3D hand keypoints, depth maps, hand masks, interacted object masks and sub-task segmentations that are all generated via our automated processing and annotation pipeline. The comprehensive annotations make EgoLive suitable not only for egocentric perception, but also for manipulation grounding, workflow decomposition and downstream human-robot transfer tasks in real-world scenarios. The detailed comparison with other egocentric datasets are listed in Tab. \ref{tab:egocentric_dataset_comparison}.

\subsection{Learning from Egocentric Data}

Human egocentric data provides visual observations and human motion data for representation and policy learning in robotic manipulation. Existing works on learning from egocentric data can be organized into three categories: \textbf{capability learning from human egocentric data}, \textbf{transfer from human data to robots}, and \textbf{extensions to whole body control}.

\textbf{Capability Learning from Human Egocentric Data} explores what capabilities for robot manipulation can be learned from human egocentric demonstrations and how larger-scale pre-training enhances such capabilities. EgoDex~\cite{Egodex}, EgoMimic~\cite{Egomimic} and EgoVLA~\cite{Egovla} investigate action-level capability learning from human egocentric data through hand trajectory imitation, co-training with human and robot demonstrations, and action-aware visual representation learning, respectively. VITRA~\cite{Vitra} and EgoScale~\cite{Egoscale} further extend this line to larger-scale pre-training by constructing robot aligned training data from human egocentric videos and studying the scaling behavior of egocentric pre-training. Together, these works show that human egocentric data is a scalable source of action representations and manipulation priors for robot learning.

\textbf{Human-to-robot transfer} studies how capabilities learned from human demonstrations can be transferred effectively to robot execution despite the embodiment gap. Phantom~\cite{Phantom}, Masquerade~\cite{Masquerade}, and MimicDreamer~\cite{Mimicdreamer} reduce the embodiment gap via image distribution alignment and demonstrate the effectiveness of the resulting supervision signal for robot policy learning. Humanoid Policy $\sim$ Human Policy~\cite{Humanoidpolicy} and Being-H0.5~\cite{Beingh05} construct shared action interfaces between human demonstrations and robot control through differentiable retargeting and semantically aligned action spaces. H-RDT~\cite{Hrdt} and EgoBridge~\cite{Egobridge} further improve transfer through staged adaptation and representation alignment, respectively. Collectively, these works improve the transfer of manipulation policies from human egocentric data to robotic systems.

\begin{figure}[t]
    \centering
    \includegraphics[width=0.93\linewidth]{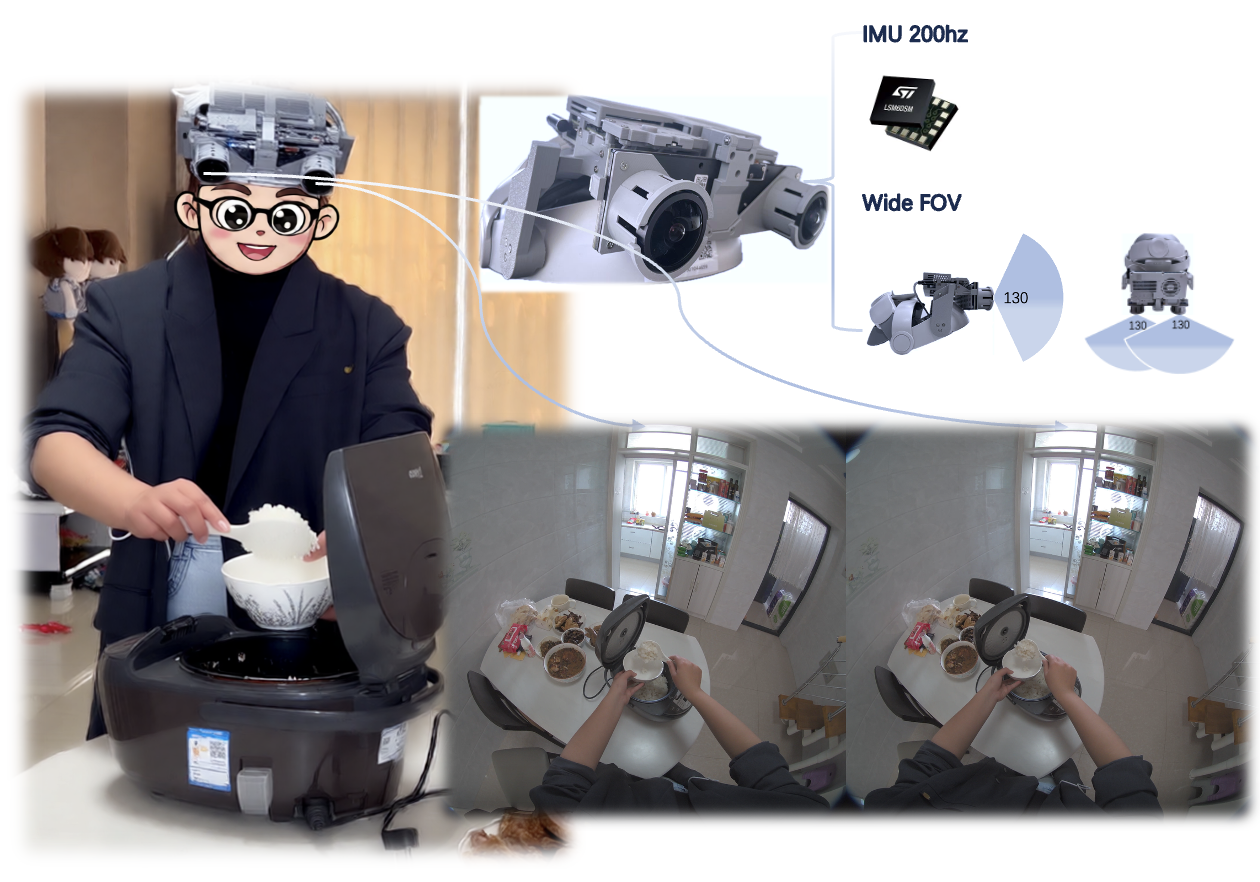}
    \caption{\textbf{Human Data Capture System} 
  The system leverages JoyEgoCam which is a custom-designed head-mounted device for acquiring human behavior data in real-world environments. It features stereo RGB cameras that provide a wide field-of-view and an integrated IMU with 200Hz measurements.
}
    \label{fig:device}
    \vspace{-10pt}
\end{figure}

\textbf{Extensions to Whole-Body Control} consider learning from egocentric data beyond hand centered manipulation to more coordinated embodied behaviors. ZeroWBC~\cite{Zerowbc}, EgoMI~\cite{Egomi}, and EgoHumanoid~\cite{Egohumanoid} extend from manipulation to humanoid whole body control, coordinated active vision and manipulation, and whole body loco-manipulation, respectively. These studies suggest that egocentric data can support not only manipulation learning, but also more general embodied control involving active vision and whole body motion.

\section{Dataset}
\label{sec:method}

\begin{figure}[ht]
    \centering
    \includegraphics[width=0.99\linewidth]{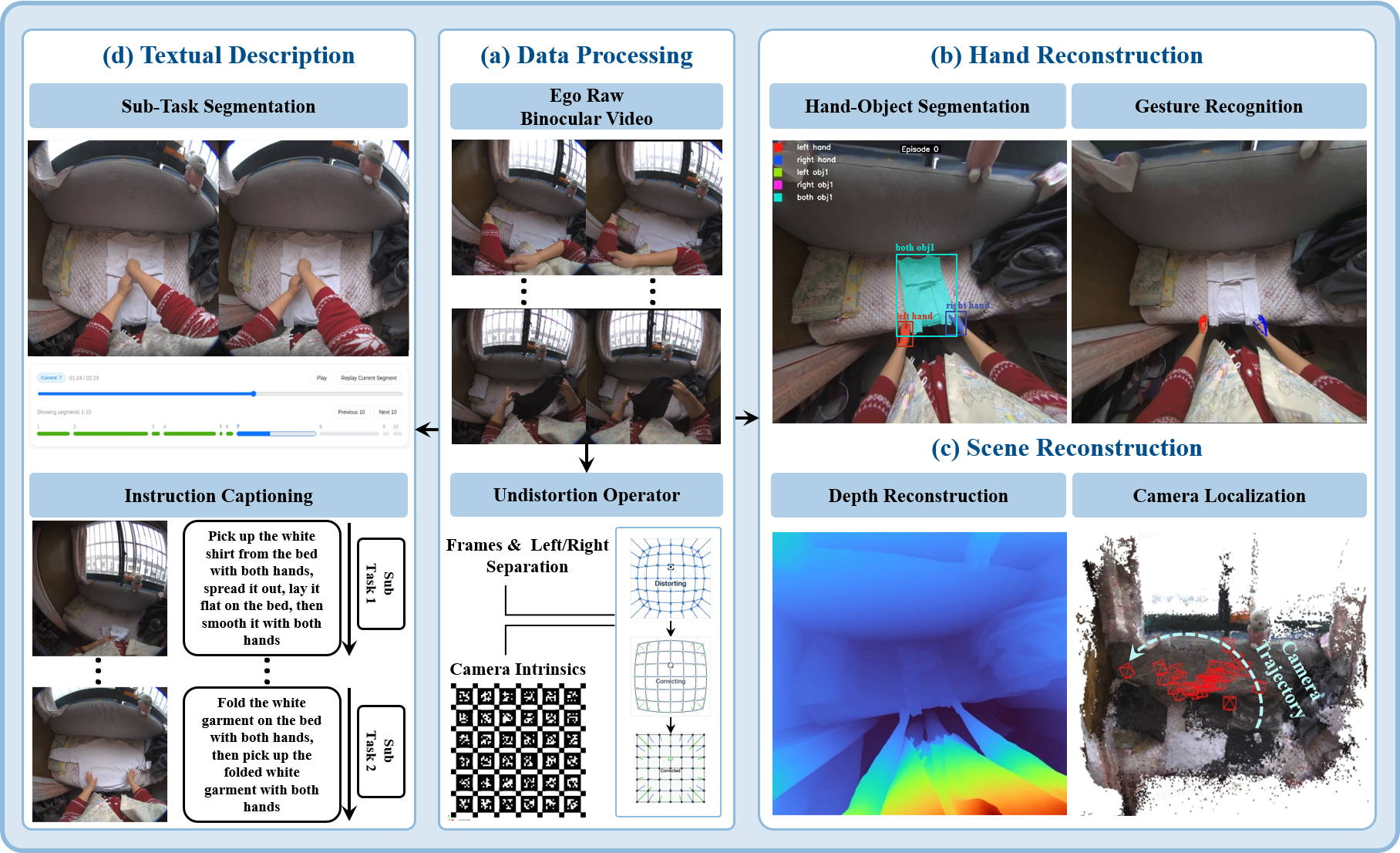}
    \caption{Overview of the automated annotation pipeline for egocentric binocular videos.
(a) The raw egocentric binocular video is first undistorted, with left/right view separation and camera intrinsic calibration.
(b) Hand reconstruction is conducted, integrating semantic gesture recognition and hand-object segmentation to capture interaction dynamics.
(c) Dense depth reconstruction and camera localization are performed to recover the scene structure and estimate the camera trajectory based on the rectified video and sensor cues.
(d) In parallel, sub-task segmentation and instruction captioning are employed to generate temporal textual description.}
    \label{fig:pipeline}
    \vspace{-10pt}
\end{figure}

\textbf{EgoLive} is a large-scale in-the-wild egocentric manipulation dataset with rich and high-quality annotations. To enable massive-scale data collection, we introduce a lightweight ergonomic headset that allows for long-term in-the-wild acquisition. We further develop an automated annotation pipeline to generate comprehensive, high-quality annotations covering action trajectories, semantic understanding and 3D reconstruction. Extensive statistical analysis demonstrates that EgoLive has a wide coverage and high local density, exhibiting rich data diversity.

\subsection{Data Collection}

All data is collected using \textbf{JoyEgoCam} (see Fig. \ref{fig:device}), a custom-designed head-mounted device for human behavior acquisition in the wild. JoyEgoCam is equipped with stereo RGB cameras that provide a wide FoV analogous to human binocular vision. The high resolution and frame rate facilitate fine-grained and agile human motion tracking, while the integrated Inertial Measurement Unit (IMU) enables improved accuracy in camera pose estimation.

The data is primarily collected in real-world scenes by recruited operators. In contrast to VR headsets that occlude users' faces and wearable devices that interfere hand motion, our device adheres to a minimal-intrusion design principle. Consequently, users experience little discomfort and can perform daily actions naturally during collection. Benefiting from the head-mounted form factor and human-like field of view, natural hand movements can be captured realistically.

The raw multi-modal data comprises stereo RGB videos with $2160 \times 2160$ resolution at $60$Hz, paired with camera calibration files, camera trigger frame timestamps, and time-synchronized IMU measurements at $200$Hz. 
Each video typically spans $1$ to $3$ minutes, featuring a single continuous mid- to long-duration manipulation activity performed in real-world environments.

\subsection{Automated Data Annotation}

As illustrated in Fig. \ref{fig:pipeline}, we develop a comprehensive in-house data annotation pipeline for frame-wise processing of multi-modal data streams. The pipeline integrates three key components:

\textbf{Motion Tracking}: 6D trajectories of both wrist and hand joints are estimated and then synchronized with camera ego-motion estimation to establish action reference frames. The hand motion is estimated by a two-stage method based on HaMeR \cite{pavlakos2024reconstructing}. MANO \cite{MANO} parameters are firstly estimated from each monocular video, which ensures precise 2D projections. The camera's ego-motion is estimated by ORB-SLAM3 \cite{ORBSLAM3_TRO2021} which fuses binocular RGB images and IMU data, with initialization starting at the first frame of the left camera.

\textbf{Semantic Understanding}: Hierarchical semantic information, including hand-object interaction states and natural language descriptions of actions, is obtained via a data processing pipeline integrating detection, tracking, segmentation and LLM-driven textual annotation techniques. Human hands and corresponding interacted objects are detected using the model proposed in \cite{Shan20} and the outputs are tracked via BoT-SORT \cite{aharon2022bot}. SAM2 \cite{ravi2024sam2} is adopted to generate segmentation masks for hands and interacted objects. Each episode is partitioned into sub-tasks according to hand-object detection and tracking results. A fine-tuned Qwen3-VL-32B \cite{Qwen3} model then takes these segmented sub-task clips as input and employs a multi-stage reasoning strategy at inference time to generate fine-grained descriptions.

\textbf{3D Reconstruction}: Accurate 3D hand reconstruction and depth estimation can be achieved by effectively leveraging binocular vision. In the second stage of hand motion estimation, stereo optimization is performed based on the previously estimated MANO parameters, yielding consistent 3D hand keypoint positions that adhere to the 2D projection constraints within each monocular video. Depth is reconstructed with $1152 \times 1152$ resolution from finely calibrated stereo RGB videos using FoundationStereo \cite{wen2025stereo}.

Overall, this pipeline enables streamlined acquisition of geometric, kinematic, and semantic annotations from stereo RGB and IMU inputs, providing multi-modal supervision signals for downstream tasks such as egocentric perception, hand-object interaction understanding and manipulation policy learning.

\begin{figure}[!t]
\centering
\includegraphics[width=0.9\linewidth]{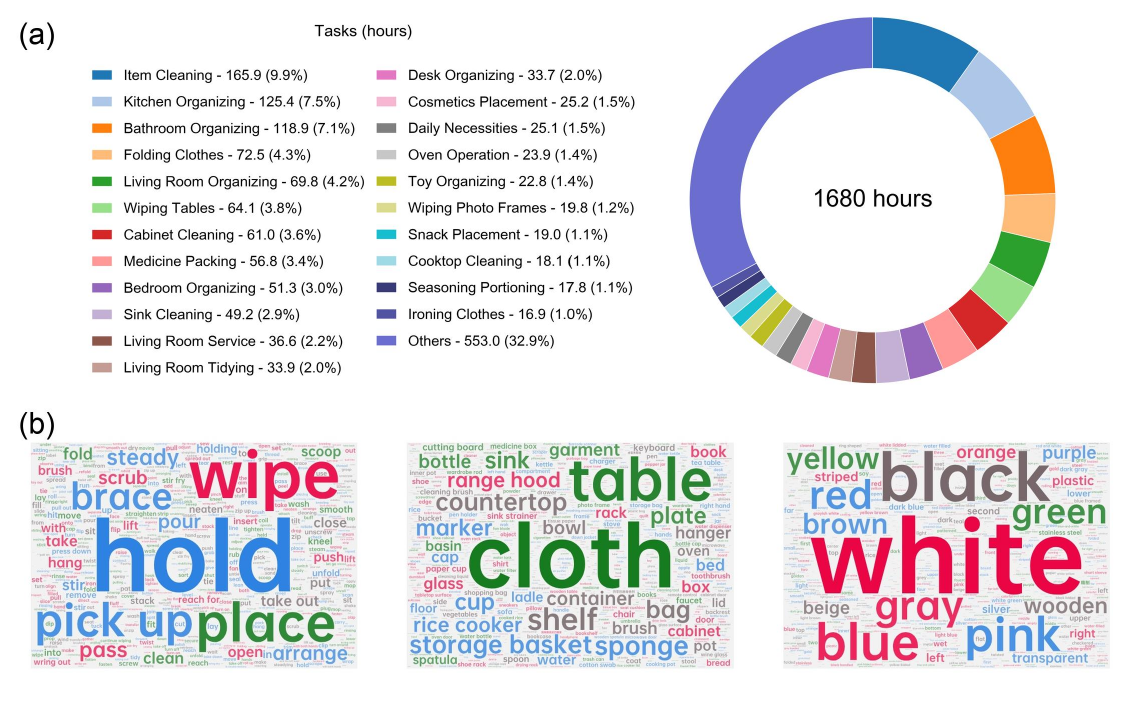}
\vspace{-0.2in}
\caption{
Discrete semantic composition of EgoLive derived from task description.
(a) Distribution of task categories, illustrating coverage of real-world activity domains including household services, organization, cleaning, logistics, and other manipulation-intensive scenarios.
(b) Word clouds of high-frequency semantic labels extracted from instruction captions across action, object and object attribute.
}
\label{fig:dataset_distribution}
\end{figure}

\subsection{Dataset Distribution}

Prior work on foundation models has shown that data diversity is critical for robust generalization \cite{brown2020gpt3,radford2021clip,chen2023palix,schuhmann2022laion5b,gadre2023datacomp}. Similar observations have also been made in robot learning \cite{openx2023rtx,Gen-0,Gen-1}, motivating work on data-centric strategies such as scaling, deduplication, re-weighting, and quality-aware data selection to refine training data distributions and thereby to improve generalization performance of embodied models  \cite{Egoscale, ReMix, zhang2026scizor}. In this section we present a systematic analysis of EgoLive from two complementary perspectives, discrete and continuous, to show that Egolive achieves improved data diversity in terms of both coverage and density in comparison to existing egocentric datasets.

\begin{figure}[h]
\centering
\includegraphics[width=0.9\linewidth]{figs/word_frequency_threshold_curves.pdf}
\caption{
Distributions of semantic labels derived from instruction captioning across datasets:
(a) object distribution,
(b) action distribution,
(c) attribute distribution.
The $x$-axis represents the word frequency threshold $n$ and the $y$-axis shows the number of distinct words with frequency greater than $n$ ($\log$-$\log$ scale).
}
\label{fig:long_tail_distribution}
\end{figure}

\textbf{Discrete Semantic Distribution}: 

In terms of discrete semantic analysis, we extract scene categories and object–action–attribute labels from instruction captions and use their frequency statistics to compare semantic coverage and long-tailed structure across datasets \cite{Vitra}.

As illustrated in Fig.~\ref{fig:dataset_distribution}, EgoLive spans a wide spectrum of tasks, showing broad coverage of manipulation-intensive scenarios. The word clouds further demonstrate that frequent tokens cover a rich variety of actions, objects and object attributes.

Fig.~\ref{fig:long_tail_distribution} compares EgoLive with EgoDex (829 hours) \cite{Egodex} and Xperience-10M (1059 hours available to date) \cite{xperience_10m} based on word (label) frequency. It is clear that EgoLive lies above EgoDex and Xperience-10M and shows longer tails across all semantic dimensions. 
Overall, EgoLive exhibits broader semantic coverage and a more natural long-tailed distribution over objects, actions, and attributes, reflecting its superior semantic diversity compared to existing egocentric datasets.

\textbf{Continuous Feature Space Analysis}: 

In addition to discrete semantic analysis, we analyze dataset distributions in a continuous feature space. Specifically, image embeddings are extracted using Cosmos-Embed1-448p \cite{cosmos_embed} and visualized via t-SNE. A joint representation of objects, environments and actions is formed to characterize the overall structure of the egocentric data.

The local neighborhoods in the t-SNE graph reflect semantic similarity, whereas clusters correspond to consistent interaction patterns that share similar objects, actions, and scene structures. The relative distances between clusters indicate variations across different interaction patterns.

\begin{figure}[htbp]
    \centering
    \includegraphics[width=0.885 \linewidth]{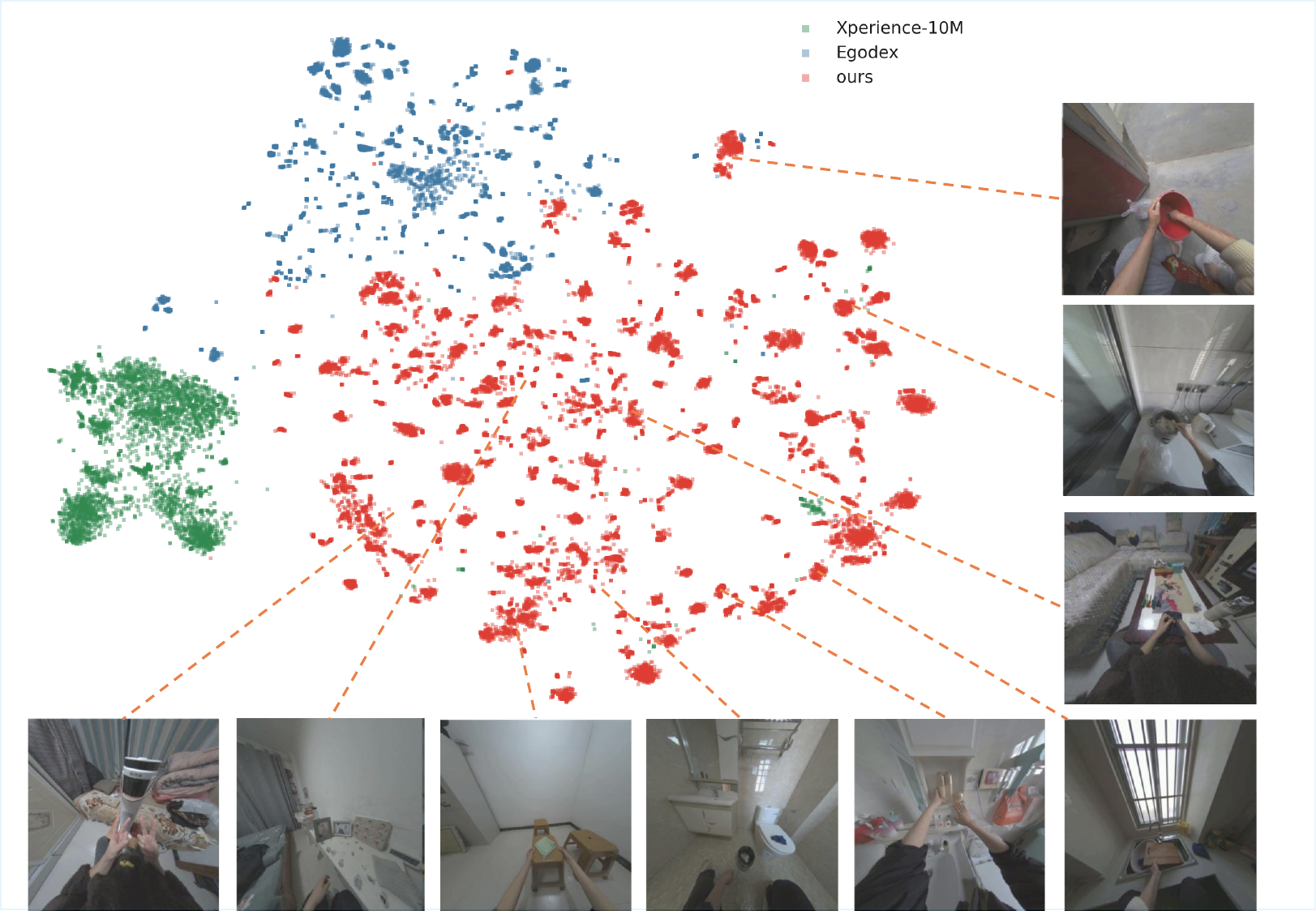}
    \caption{
    t-SNE visualization of the joint embeddings of objects, environments and actions. Example frames are representative samples from the different regions that illustrate the wide diversity of our data.
    }
    \label{fig:representation_manifold}
\end{figure}

As shown in Fig.~\ref{fig:representation_manifold}, EgoLive occupies a broader region of the representation manifold and exhibits more locally coherent clusters than EgoDex and Xperience-10M. This demonstrates that EgoLive not only achieves wider interaction patterns but also preserves clear local structures in the continuous representation space.

\begin{figure}[b!]
    \centering
    \makebox[\linewidth][l]{%
        \hspace*{-0.1cm}
        \includegraphics[width=0.99\linewidth]{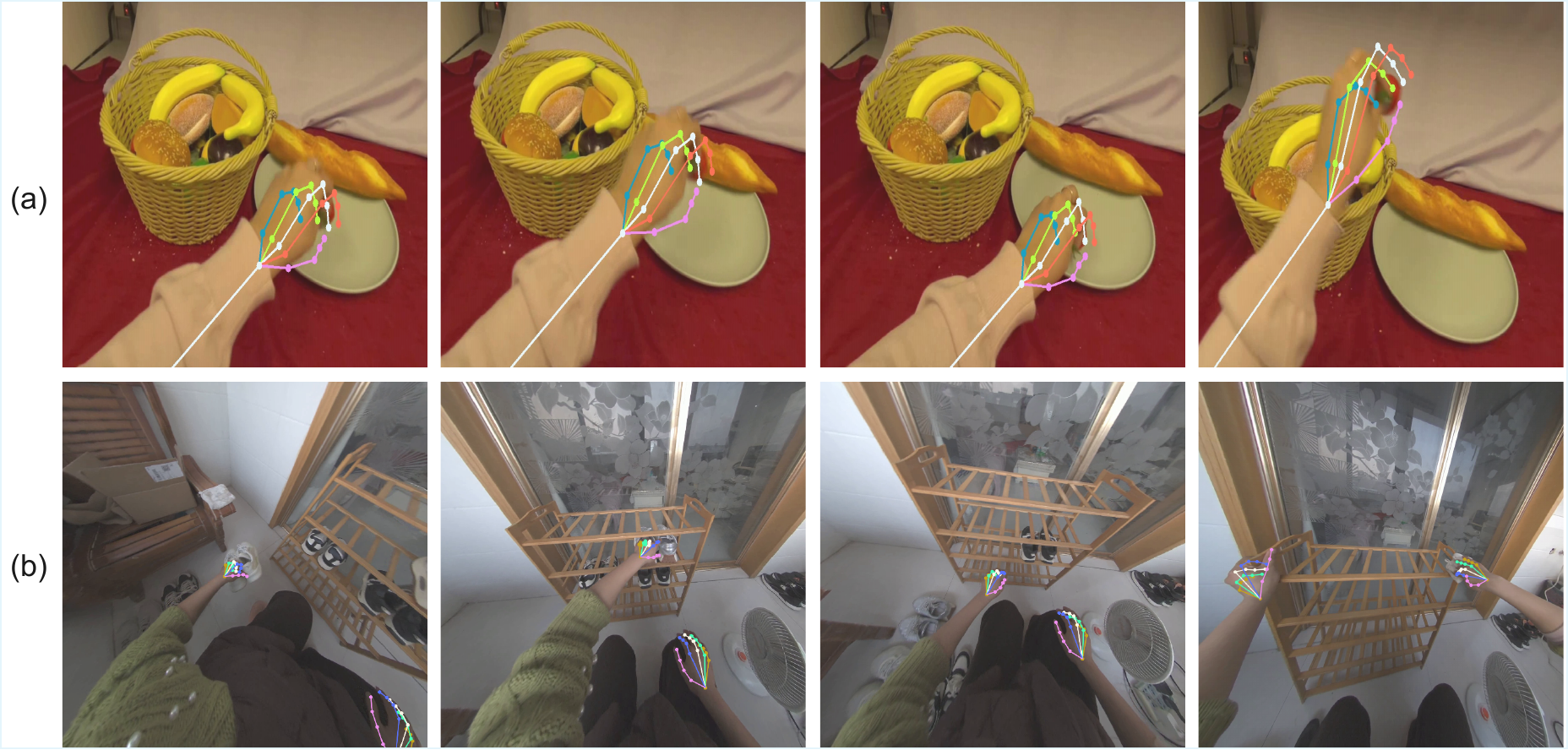}
    }
    \caption{\textbf{Qualitative Comparison of 2D Keypoint Annotations.} Visualizations of 2D keypoint annotations from (a) EgoDex and (b) EgoLive. EgoDex has non-negligible annotation errors and spatial misalignments. In contrast, EgoLive provides highly accurate and robust 2D keypoint annotations.}
    \label{fig:dual}
\end{figure}

\begin{figure}[h!]
    \centering
    \includegraphics[width=0.93\linewidth]{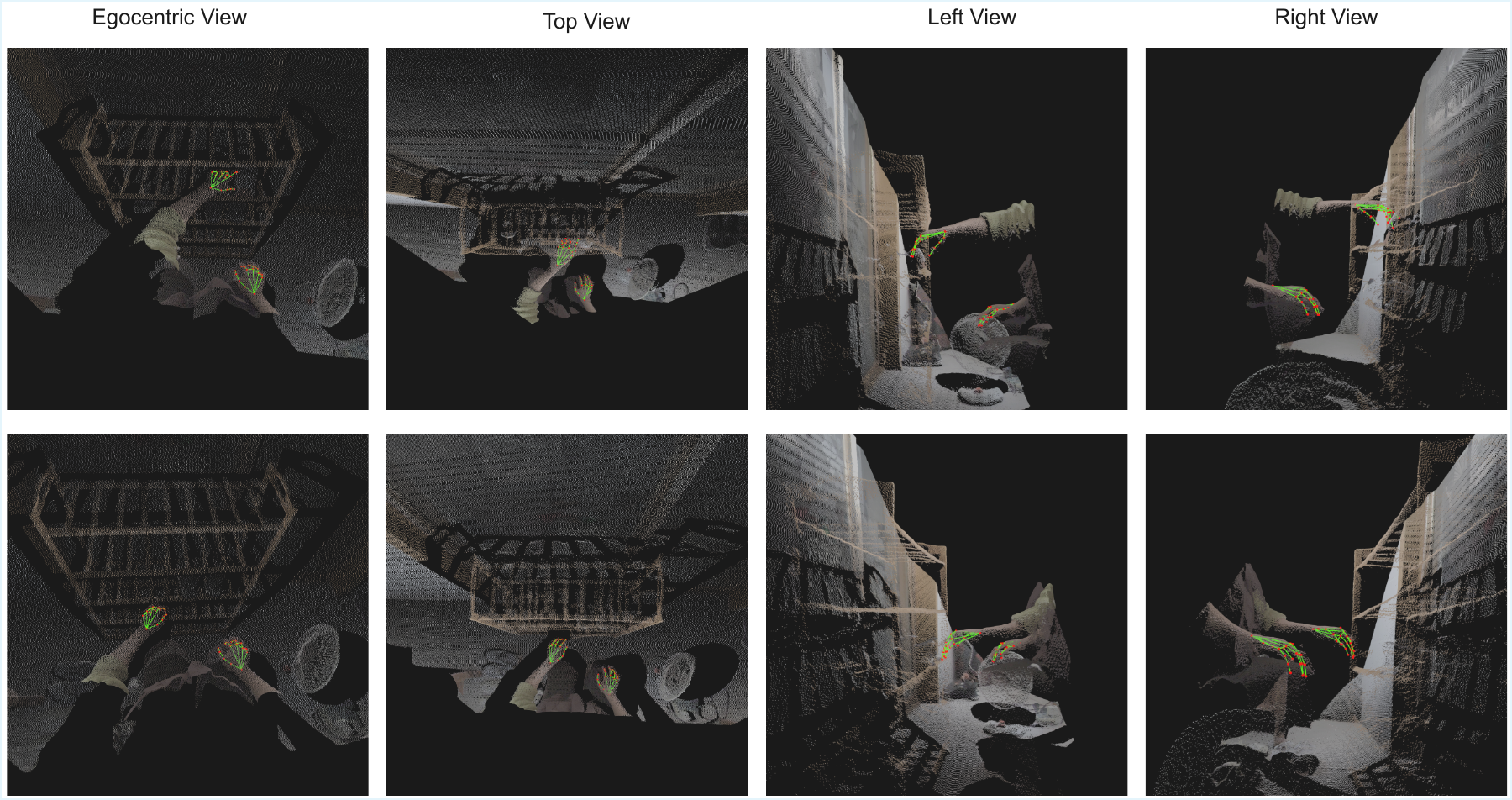}
    \caption{\textbf{Multi-view Visualization of 3D Keypoints.} 3D keypoint estimations rendered from four distinct perspectives: egocentric, front, left-side, and right-side views. Our method preserves strong spatial consistency with no wrist drift across all viewing angles.}
    \label{fig:multi-view}
\end{figure}

\section{Accuracy Evaluation} \label{sec:eval}

\subsection{Hand Reconstruction} 

To evaluate the accuracy of hand reconstruction, we first provide visual comparisons of 2D hand keypoint annotations between EgoDex and our dataset, as shown in Fig.~\ref{fig:dual}. Keypoints derived from EgoDex display obvious localization errors, resulting in significant misalignment between the projected skeleton and the actual hands in the images. In comparison our dataset generates much more accurate keypoints that align closely with hands across all frames.

In addition, our method achieves precise 3D keypoints annotations through joint optimization in the stereo space. The 3D estimation accuracy is validated by projecting the estimated keypoints onto the corresponding scene reconstructed from depth images and evaluating their alignment.
As shown in Fig. \ref{fig:multi-view}, the predicted 3D hand skeleton demonstrates excellent alignment with the point cloud representation of the hands. Building on the depth reconstruction evaluation detailed in Section \ref{sec:depth}, which achieves millimeter-level depth accuracy, we conclude that our 3D keypoint estimation achieves competitive precision. This result confirms that our hand reconstruction framework effectively addresses depth drift and maintains consistent absolute physical scale, even under severe dynamic occlusions.

\begin{figure}[htb]
    \centering
    \includegraphics[width=0.8\linewidth]{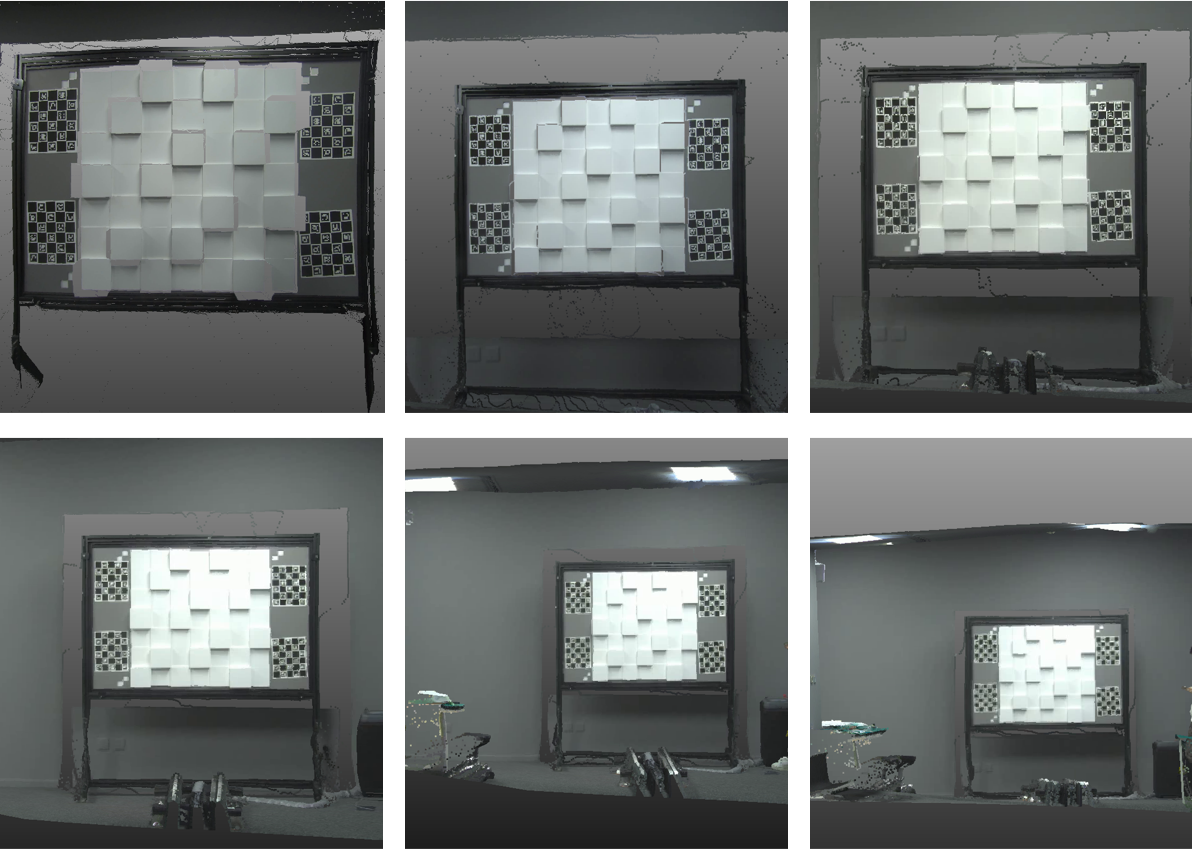}
    \caption{Point cloud reconstructions of the calibration board at distances ranging from $500mm$ to $3500mm$.}
    \label{fig:biaodingjian_visualization}
\end{figure}

\begin{table}[b!]
    \centering
    \renewcommand{\arraystretch}{1.12}
    \resizebox{\textwidth}{!}{
    \begin{tabular}{ccccccc}
        \toprule
        \textbf{Distance (mm)}& \textbf{Mean Error (mm)}& \textbf{Error $<5$ mm \%}& \textbf{Error $<10$ mm \%}& \textbf{Error $<20$ mm \%}& \textbf{Error $<30$ mm \%}& \textbf{Error $<40$ mm \%}\\
        \midrule
        500  & 3.059  & 79.576 & 99.19  & 100.0  & 100.0  & 100.0  \\
        700  & 3.045  & 79.755 & 99.833 & 100.0  & 100.0  & 100.0  \\
        900  & 5.381  & 56.86  & 87.051 & 96.987 & 99.917 & 100.0  \\
        1100 & 7.091  & 43.84  & 74.817 & 96.099 & 99.417 & 100.0  \\
        1300 & 7.033  & 48.989 & 73.227 & 95.577 & 99.697 & 100.0  \\
        1500 & 8.751  & 35.636 & 63.664 & 92.528 & 99.333 & 99.917 \\
        1700 & 13.114 & 24.233 & 43.992 & 76.874 & 93.943 & 99.079 \\
        1900 & 12.083 & 27.696 & 49.601 & 79.745 & 94.904 & 99.074 \\
        2100 & 13.014 & 26.505 & 47.433 & 74.121 & 92.053 & 99.75  \\
        2300 & 15.095 & 20.878 & 40.893 & 69.96  & 87.822 & 96.933 \\
        2500 & 13.994 & 25.061 & 47.391 & 71.72  & 88.6   & 97.527 \\
        2700 & 14.82  & 23.789 & 41.761 & 72.063 & 87.318 & 96.4   \\
        2900 & 13.414 & 31.071 & 49.34  & 71.32  & 89.229 & 98.794 \\
        3100 & 15.254 & 26.29  & 45.319 & 65.63  & 83.568 & 97.064 \\
        3300 & 17.372 & 18.98  & 36.162 & 64.54  & 81.561 & 92.889 \\
        3500 & 18.035 & 20.952 & 36.861 & 59.798 & 79.431 & 93.478 \\
        \bottomrule
    \end{tabular}
    }
    \caption{Quantitative analysis of the proposed depth reconstruction model in the calibration room. We report the mean reconstruction error and error distribution under different thresholds, evaluated at operating distances from $500mm$ to $3500mm$.}
    \label{tab:distance_error_evaluation}
\end{table}

\subsection{Depth Reconstruction}
\label{sec:depth}

We adopt a calibrated checkerboard-based method to quantitatively evaluate depth reconstruction accuracy.
Specifically, a stepped array with known depth gradients and a ChArUco calibration board are used as reference targets and captured by our system at multiple distances from $0.5m$ to $3.5m$. Following stereo rectification, the 6D poses of camera relative to the checkerboard are estimated by jointly optimizing the global reprojection error using the BFGS algorithm. Subsequently, the depth of each corner point is calculated based on the known geometry of the checkerboard, serving as the predicted ground truth. By comparing this ground truth depth with the depth predicted from our depth reconstruction method, as illustrated in Fig. \ref{fig:biaodingjian_visualization}, we obtain the quantitative evaluation results of the model, which are presented in Tab. \ref{tab:distance_error_evaluation}. These results demonstrate that our depth reconstruction  approach can achieve low mean errors within the typical human operation range (approximately less than $0.9m$).

\begin{figure}[h!]
    \centering
    \includegraphics[width=0.8\linewidth]{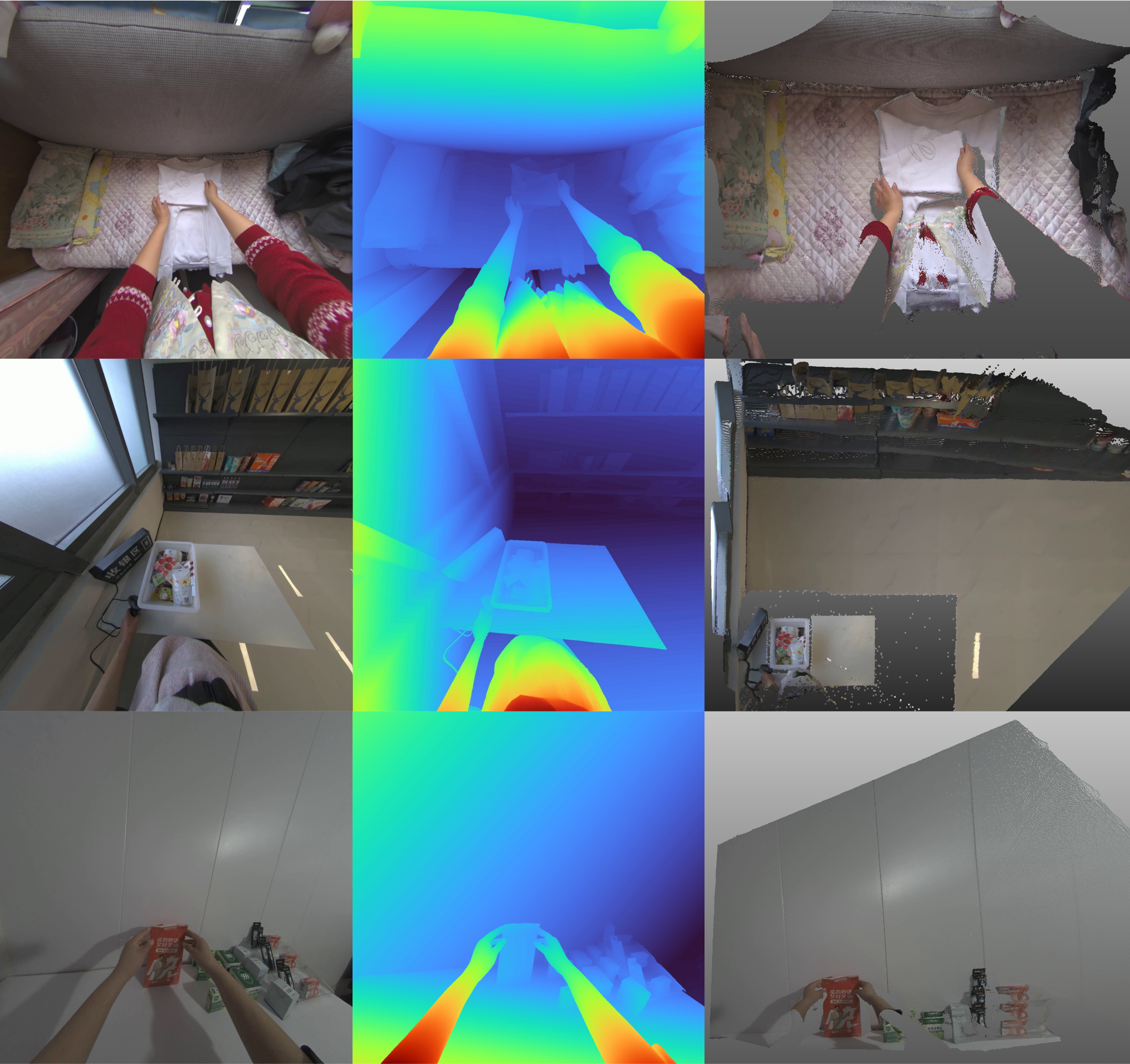}
    \caption{Visualization of depth and point-cloud reconstruction results in real-world scenes from EgoLive dataset.
    From left to right at each row: original left image, predicted depth map and reconstructed point clouds. The spatial structure and 3D geometric information across diverse real-world scenes can be recovered effectively by our proposed method.}
    \label{fig:real_scenes_visualization}
\end{figure}

To further assess the qualitative performance and practical applicability of our proposed method, three representative collection scenes are considered in Fig. \ref{fig:real_scenes_visualization}. The visual results demonstrate that JoyEgoCam, together with the depth reconstruction method \cite{wen2025stereo}, enables high-quality depth estimation that faithfully recovers the underlying spatial structures and complex 3D geometric information.

\subsection{Instruction Captioning}
\begin{figure}[h]
\centering
\includegraphics[width=0.99\textwidth]{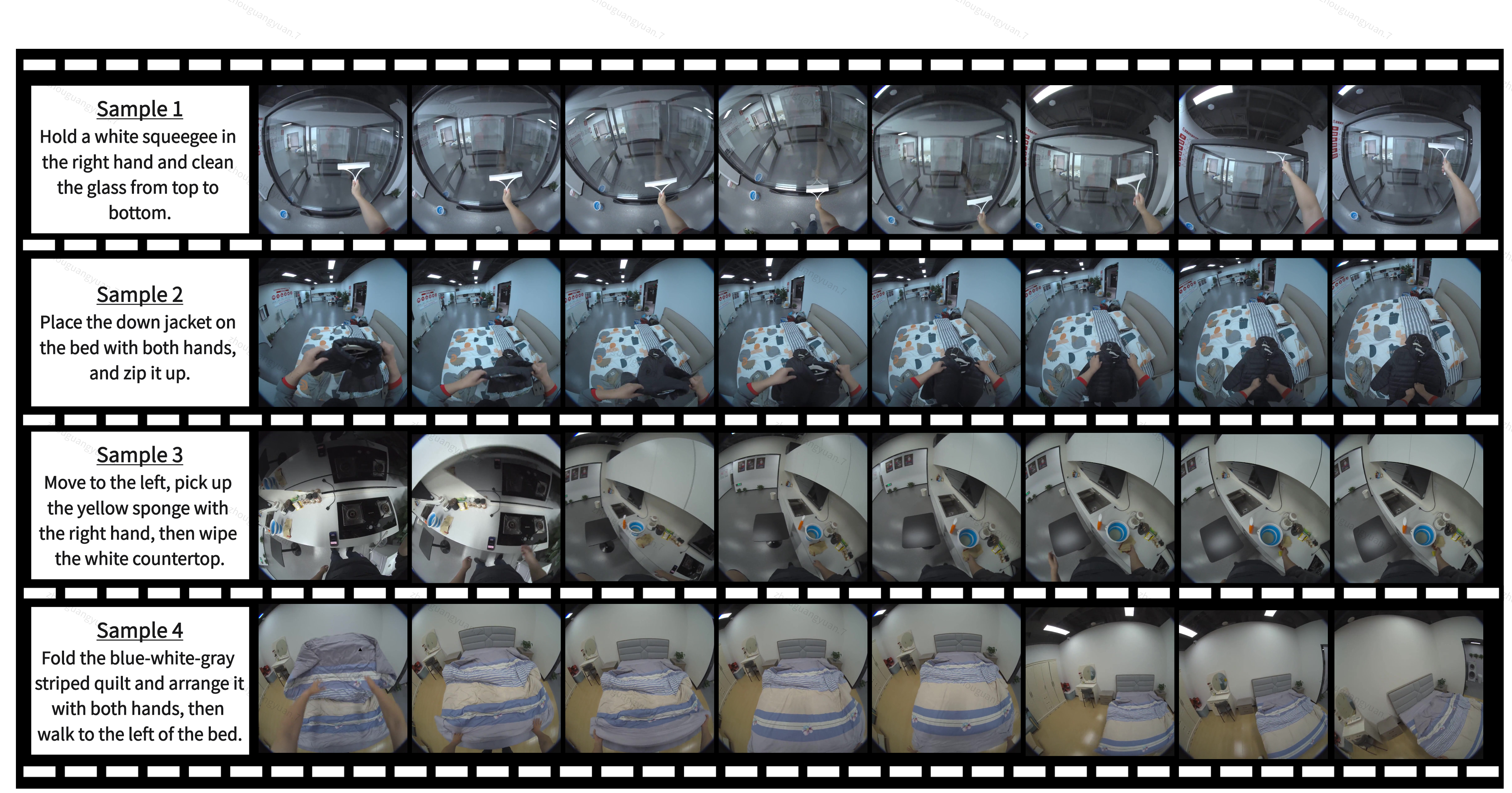}
\caption{Qualitative examples from our instruction-based annotation pipeline across diverse scenes, including glass wiping, clothes folding, refrigerator cleaning, and bed making.}
\label{fig:Caption}
\end{figure}

For each episode, the video is segmented into sub-task clips corresponding to complete atomic actions with durations ranging from approximately $1$ to $20$ seconds. An instruction captioning module is introduced to generate structured semantic annotations for each clip. Instead of producing completely free-form descriptions, the module explicitly models three essential elements for understanding egocentric manipulation: \textbf{hands}, \textbf{manipulated objects} and \textbf{actions}. This design is motivated by the observation \cite{Vitra, egocentric-hoi, Beingh0} that sub-tasks are tightly coupled with hand action and the manipulated object in human video captioning. As shown in the top row of Fig.~\ref{fig:Caption}, a naive natural language description may be $<$\textit{use a squeegee to wipe the glass door}$>$ whereas our description provides fine-grained information about hand usage, manipulated objects, and action. 

\begin{table}[b!]
\centering
\scriptsize
\setlength{\tabcolsep}{3pt}
\renewcommand{\arraystretch}{1.15}
\begin{tabular}{c >{\raggedright\arraybackslash}p{4.6cm} >{\raggedright\arraybackslash}p{4.2cm} cccc}
\toprule
Sample & \multicolumn{1}{c}{GT Caption} & \multicolumn{1}{c}{Predicted Caption} & Action & Object & Hand & Global \\
\midrule

1 &
Clean the glass from top to bottom with the white squeegee in the right hand. &
Hold a white squeegee in the right hand and clean the glass from top to bottom. &
$\checkmark$ & $\checkmark$ & $\checkmark$ & $\checkmark$ \\ \hline

2 &
Place the black cotton coat hanging on the hanger flat on the bed, then zip it up with both hands. &
Place the down jacket on the bed with both hands, and zip it up. &
$\checkmark$ & $\checkmark$ & $\checkmark$ & $\checkmark$ \\ \hline

3 &
Remove the cloth covering the sponge with the right hand. Then pick up the sponge with the right hand. &
Move to the left, pick up the yellow sponge with the right hand, then wipe the white countertop. &
$\times$ & $\times$ & $\checkmark$ & $\times$ \\ \hline

4 &
Spread and smooth the blue-and-white striped duvet flat on the bed with both hands. Then grasp its two corners with both hands, lift it, and fully unfold it. Finally, walk toward the left of the bed. &
Fold the blue-white-gray striped quilt and arrange it with both hands, then walk to the left side of the bed. &
$\times$ & $\checkmark$ & $\checkmark$ & $\times$ \\

\bottomrule
\end{tabular}
\caption{Detailed evaluation results for the examples in Fig.~\ref{fig:Caption}.}
\label{tab:caption}
\end{table}

To evaluate the quality of captioning results, we adopt a granular caption assessment protocol using LLM-as-a-Judge that assesses the consistency of key elements with ground-truth descriptions, including hand consistency, object consistency, action consistency, and global consistency. Hand consistency measures whether the caption correctly identifies the interaction hand and object consistency evaluates whether the manipulated object matches the visual content, including object category, color, and shape. Action consistency evaluates whether the described manipulation action accurately reflects the observed behavior. Global consistency measures whether the caption provides a faithful and coherent description of the sub-task clip, taking into account the joint correctness of hand usage, object interaction, action semantics as well as sentence naturalness and completeness. We present corresponding consistency evaluation examples in Tab. \ref{tab:caption} for the cases shown in Fig. \ref{fig:Caption}.
\section{Conclusion}
This work introduces \textbf{EgoLive}, the world's largest open-source egocentric dataset dedicated to human daily operational activities within realistic, utility-driven working scenarios. Leveraging purpose-built hardware, a scalable collection pipeline and a high-accuracy annotation framework, EgoLive delivers industrial-grade high-quality human demonstration data suitable for robot learning. As a substantial addition to existing public egocentric datasets, it offers qualitative improvements over prior work in data scale, data quality and diversity. More importantly, building upon deployment-friendly collection devices and a streamlined data production pipeline, EgoLive is designed to sustain continuous growth in both scale and coverage, establishing an ever-growing knowledge base of natural human behavioral priors. It is poised to inspire and support key research directions in the robotics community, such as human-to-robot alignment, humanoid robot policy learning, and ultimately to bridge the gap between human behavior and robot action. 

\section{Contributions}

\textbf{Contributors}

Yihang Li\textsuperscript{\dagger}, Xuelong Wei, Jingzhou Luo, Yingjing Xiao, Yibo Bai, Guangyuan Zhou, Teng Zou, Chenguang Gui, Jiajun Wen, He Zhang,  Kangliang Chen, Xing Pan, Shuaiyan Liu, Daming Wang, Tao An, Jiayi Li, Shibo Jin, Wanwan Zhang, Tianyu Wang, Boren Wei, Zhixuan Huang, Fangsheng Liu, Ruodai Li, Hui Zhang, Anson Li, Yicheng Gong, Peng Cao, Jiaming Liang, Liang Lin\textsuperscript{\dagger}.

\renewcommand{\thefootnote}{\fnsymbol{footnote}}
\footnotetext{\textsuperscript{\dagger}Corresponding author: Yihang Li <liyihang18@jd.com>, Liang Lin <linliang@ieee.org>}

\clearpage
\bibliographystyle{plainnat}
\bibliography{cite}

\end{document}